# Temporal Video-Language Alignment Network for Reward Shaping in Reinforcement Learning

Ziyuan Cao

Reshma Anugundanahalli Ramachandra

Kelin Yu

coliny@gatech.edu

*Abstract—* Designing appropriate reward functions for Reinforcement Learning (RL) approaches has been a significant problem, especially for complex environments such as Atari games. Utilizing natural language instructions to provide intermediate rewards to RL agents in a process known as reward shaping can help the agent in reaching the goal state faster. In this work, we propose a natural language-based reward shaping approach that maps trajectories from the Montezuma's Revenge game environment to corresponding natural language instructions using an extension of the LanguagE-Action Reward Network (LEARN) framework. These trajectory-language mappings are further used to generate intermediate rewards which are integrated into reward functions that can be utilized to learn an optimal policy for any standard RL algorithms. For a set of 15 tasks from Atari's Montezuma's Revenge game, the Ext-LEARN approach leads to the successful completion of tasks more often on average than the reward shaping approach that uses the LEARN framework and performs even better than the reward shaping framework without natural language-based rewards.

## I. INTRODUCTION

Reinforcement Learning (RL) has been used extensively in games with promising results in terms of performance. RL algorithms utilize the concepts of rewards (positive or negative reinforcements) to help agents in learning to optimize their behaviors in an environment with which they interact. Subsequently, RL has been successfully implemented in complex environments such as Atari games [1]. However, defining reward functions has been the most significant challenge in RL, especially while scaling RL algorithms for real-world applications with large state spaces. Rewards must be defined accurately and efficiently to clearly represent the tasks to be completed by the agent. There have been several attempts to combat the problem of defining reward functions, one of which is to define sparse rewards such as providing positive rewards for reaching the goal state while the agent receives no rewards if it fails to reach the goal state [2]. This could be an extreme solution that would lead to slow and difficult learning. On the other hand, rewards could be specified in a dense manner such as the running score in an Atari game [3]. Although such rewards are easier to learn from since the agent can randomly take series of actions to easily find rewards, it would be difficult and time-consuming to specify rewards in such detail. Thus, specifying intermediate rewards designed to help the agent in reaching the goal state could be a solution to defining rewards. This process is known as reward shaping [4].

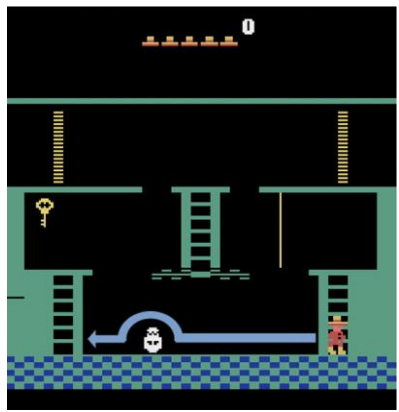

Fig. 1 An RL agent in the Montezuma's Revenge Atari game environment meant to follow the blue trajectory in order to reach the golden key.

Fig. 1 shows an instance of an RL agent interacting with the complex Atari game environment, Montezuma's Revenge. If the goal is to obtain the golden key, the intermediate steps include moving to the left, jumping over the skull, and moving towards the key. If the agent is only rewarded once it reaches the goal state, as is the case while defining sparse rewards, then the agent wastes a lot of time exploring the environment while it learns to reach the goal state. Thus, providing intermediate rewards that advance the agent towards the goal state or reward shaping could help in saving exploration time and reduce the number of agent's interactions with the environment. Various approaches have been used for rewards shaping in RL [5][6]. However, translating human inputs to real-valued rewards could be challenging due to cognitive bias and requirement of programming experts who can accurately map instructions to actions in the trajectories. Designing or accurately specifying intermediate rewards is another problem that is quite difficult to solve for people not well-versed with the gaming environment or strategies.

Natural language has been shown to be successful in communicating with agents powered by reinforcement learning and imitation learning [7]. Utilizing natural language for reward shaping helps in designing dense rewards. For tasks as seen in the example in Fig. 1, providing instructions in natural language in order to design intermediate rewards is seen to improve the performance of RL agents [8]. When the agent is given instructions such as "Jump over the skull while going to the left", learning is faster, and the agent completes the task 60% more often on average when compared to learning without using natural language-based reward shaping. Since non-experts can also provide natural language instructions, specifying rewards or describing the task environment becomes more straight-forwards and novel skills can be taught to RL agents in an easier and convenient manner. However, this approach has several drawbacks.

Approaches that exploit natural language instructions for reward shaping train a language reward model that predicts how a piece of trajectory matches with a language command using a collected dataset of <instruction, trajectory> pairs. However, this approach requires a separate supervised training phase for grounding the language commands in the environment in which the agent will be deployed. For example, the instruction "Jump over the snake" needs grounding of object "snake" to the corresponding pixels in the image of the current state and "jump" must be mapped to the appropriate action in the action space. Collection of thousands of <instruction, trajectory> pairs is needed. This makes the design hard to scale in terms of generalizing to other environments. Another drawback is that an oversimplified encoding strategy of trajectories, which ignores temporal and state information, is utilized as the input to the language reward model. A better encoder of trajectories may provide more precise reward shaping. Thus, this work proposes improvements to these previous natural language-based reward shaping techniques in order to resolve these limitations and improve natural-language-instructions-to-reward mapping and, in turn, the performance of the RL agent.

The proposed approach for natural language-based reward shaping first obtains the word embeddings of the natural language instructions corresponding to the trajectories taken from the Atari Grand Challenge dataset [9]. Bidirectional Encoder Representations from Transformers (BERT) [10] is utilized to obtain the word embeddings since it captures the underlying context of the natural language instructions, especially in cases where the given instructions are with respect to time or space. Further, the proposed approach considers the entire trajectory as complete action sequences while previous approaches aggregated the sequence of past actions into an action-frequency vector. Additionally, an alignment model is also proposed to map the entire trajectory frames to the corresponding language descriptions. This helps in comparing and validating the trajectory and corresponding language descriptions generated before using them to design the reward function.

The main contributions of this work include:
i. Generation of natural language instructions using BERT to obtain word embeddings of the instructions.
ii. Implementation of an alignment model that maps the entire trajectory to the corresponding language description in order to discriminate between matched and unmatched trajectory-language pairs.

The rest of this paper is divided as follows: Section II includes a brief summary of the relevant related works, Section III details the proposed language-based reward shaping model and the language-aided RL model. Section IV details the experimental setup, Section V presents the results while Sections VII and VIII discuss the results and conclusions of the work respectively.

II. RELATED WORKS

The intersection of natural language with RL algorithms has been explored in multiple ways. On one hand, popular Natural Language Processing (NLP) tasks such as question-answering and text generation have been solved using RL [11][12]. In such tasks, for example, with the task of summarization, training based on supervised learning suffers from the inherent assumption that the ground truth is available for every training step otherwise known as exposure bias. Thus, combining RL's global sequence prediction training with the standard word prediction helps improve the performance on NLP tasks [13]. With respect to learning an optimal policy using RL algorithms, reward functions must be defined accurately. For environments that provide dense reward signals such as the game scores in Atari game environments, an agent policy can be trained such that it outperforms human players for most of the games [14]. However, in the environment considered in this work, games like Montezuma's Revenge provide sparse reward signals, making it difficult to learn an optimal policy. Environments with sparse reward signals also demand more data and resources [15]. Thus, reward shaping is used to augment the reward function with inputs from human observers in order to increase the number of rewards which, in turn, speeds up the policy learning process [16]. Human inputs can be given in terms of task demonstrations [17], just good or bad binary type of inputs as in TAMER [18] or through natural language [19][20]. The problem with the first two types of human feedback is the fact that it requires expert programmers who can map optimal demonstrations and also know the right action to take at every point in the game. Utilizing natural language, on the other hand, can help even non-experts provide feedback.

Using natural language-based instructions in learning RL policies has been explored quite extensively in the past. From adventure games based on language [21] to tasks involving robots avoiding obstacles or navigation [19][20], natural language has been exploited for providing instructions. [22] proposes a language-action reward network which implements a reward function by grounding language commands in the environment. Another approach included data obtained from human teachers describing in detail the tasks that they were performing [23]. The descriptions provided by the teachers were mapped to a reward function which was used to train the RL agent to learn policies. [24] is another similar approach, more in-line with our work, where natural language narration was used for reward shaping for tasks in the Montezuma's Revenge game environment. The results showed that the approach outperformed methods without reward shaping.

Several works have explored the problem of how RL agents learn to follow natural language instructions. [25] combines the state information with the natural language description of the goal to learn policies. Reward shaping is performed based on the agent's distance from reference trajectories and from the goal. Another straightforward approach to using natural language for reward shaping can be seen in [26], which describes rules using natural language to decide whether to choose an action. [27][28] also perform language-to-reward mapping by designing rewards based on if the agent arrives at the right location or not by following the instruction provided. These approaches still face the problem of efficiently defining reward functions since it requires an expert programmer who can make decisions about how language instructions are mapped to the environment.

However, all these approaches are difficult to scale to more complex environments since scaling the simple rule-based features would require a lot of engineering for different instances of the tasks. Describing instructions using natural language and mapping them directly to reward functions, as done in our approach, could be more convenient and scalable. The proposed approach solves a language conditional RL task utilizing NLP methods for assistive learning, thus combining the two ways in which RL and NLP have been combined previously [7].

Approaches such as [29][30] have also explored specifying subgoals using natural language instructions in order to improve how the RL agent follows the instruction specified using natural language. Our approach, on the other hand, focuses on improving RL algorithms using natural language instructions. [31] more recently proposes a language-conditioned offline reward learning approach that learns language-conditioned rewards effectively from annotations obtained through robot interactions with the environment. To solve the problem of grounding language in the robot's observation space, datasets obtained offline are leveraged to learn whether the language instruction is completed when there is a change in the state for offline multi-task RL. However, language-conditioned instruction-following RL agents in such approaches utilize action spaces which are simple and parameterized. Once the instructions are provided, they are fixed over the entire execution of the agent with little opportunity provided for subsequent interactions by the instructor.

Previous works have also explored utilizing features from the instruction-reward mappings in action-value functions [32]. Most similar to our work is the LEARN model introduced in [8]. LEARN or LanguagE-Action Reward Network trains a neural network to take as input a trajectory-instruction pair and produce as output the prediction probabilities of how well the language instruction describes the actions taken in the trajectory. These probabilities are used to design the reward function that is used to train the RL agent. Possible drawbacks with such an approach include temporal ordering, state-based rewards and multi-step instructions. LEARN only considers past actions while forming its action-frequency vector instead of considering the complete action sequence as is done in our approach. For example, for an instruction such as "Jump over the skull while going to the left", intermediate language-based rewards are designed by considering trajectories with higher frequencies associated with the "jump" and "left" actions. The drawback of such an approach is that it completely ignores temporal and state information. The instruction "Jump over the skull while going to the left" provides a significant amount of important information as to where the agent should jump towards and when should it make the jump. Jumping left is different from jumping towards left while avoiding the skull. Such an approach would not utilize the state information provided by instructions such as "Go towards the key" or "Avoid the pit". Evidently, the reward functions defined using the language instructions could be modelled as a function of both the past actions and states to improve the learning. Thus, our approach utilizes a contextual language-reward mapping that considers information provided by the entire instruction rather than taking actions based on only high-frequency action words.

III. PROPOSED APPROACH

*A. Dataset and Preprocessing*

The RL agent was trained on the dataset obtained from [8]. This data consists of 20 trajectories from human gameplays of the Montezuma's Revenge game obtained from the Atari Grand Challenge dataset [9]. 2,708 three-second-long frames were extracted from these trajectories to generate 6,870 language descriptions using Amazon Mechanical Turk (see supplementary material for examples of language annotations). These language annotations are instructions for the agent to follow in order to reach the goal state from the start state.

For the model to understand the natural language of annotations for each of those trajectories, BERT was used to obtain word embeddings of the language instructions. A transformer model such as BERT is used instead of pre-trained embeddings or RNN with GloVe [33] to obtain the word embeddings for these natural language instructions. BERT is able to capture the underlying context of the natural language instructions, especially in cases where the given instructions are with respect to time or space. Thus, the proposed approach considers the entire trajectory as complete action sequences while previous approaches aggregated the sequence of past actions into an action-frequency vector, thereby losing temporal and state information.

*B. Model*

We consider an augmented version of a Markov Decision Process (MDP) in RL, defined by $MDP'$: $\langle S, A, T, R, \gamma, l \rangle$ where $S$ denotes set of states, $A$ is the set of actions, $T: S \times A \times S \rightarrow [0,1]$ is the transition probabilities of going from current state $s_t \in S$ to state $s_{t+1} \in S$ by taking action $a_t \in A$. The reward function $R: S \times A \rightarrow \mathbb{R}$ maps current state-action $(s_t, a_t)$ pair to corresponding real-valued rewards, $\gamma$ is the discount factor. The extension to this MDP is provided by including the natural language instruction $l$. RL uses this $MDP'$ in order to learn an optimal policy $\pi^*$ that maximizes the expected sum of rewards. This process of learning an optimal policy is two-fold involving the Ext-LEARN framework and using the language instructions for reward shaping in RL.

The extended version of LanguagE-Action Reward Network (LEARN) uses the entire trajectory $\tau$ and the corresponding language instruction $l$ obtained from data preprocessing (Sub-section A) i.e., the pair $(\tau, l)$ as input to train a neural network to predict if the language accurately describes the actions taken in that particular trajectory. To train the Ext-LEARN model, we need both positive and negative examples of trajectory-language pairs. In a positive example, the language correctly describes the trajectory. In a negative example, the language does not describe the trajectory. The positive examples come directly from the dataset manually collected in [8]. To sample negative examples, we randomly sample a language instruction which describes a different section of the same trajectory. This results in 12K pairs consisting of the same number of positive and negative examples.

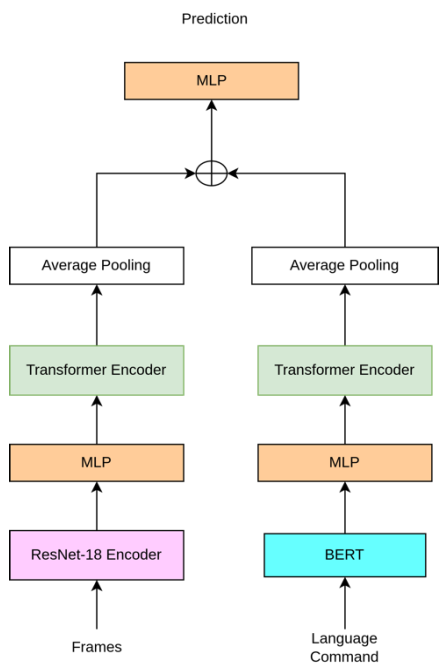

Fig. 2 Overview of proposed approach.

A diagram of the proposed Ext-LEARN model is shown in Fig 2. Ext-LEARN model takes a trajectory-language pair as the input. The trajectory is represented by a sequence of frames of the UI of the Montezuma's Revenge game. As provided in the dataset collected in [8], each trajectory consists of 150 frames. For each trajectory, we extracted 15 evenly spaced frames from the 150 frames. These extracted frames are fed into a pretrained image encoder to generate frame feature vectors. The language instruction $l$ is fed into a pretrained language encoder to generate language feature vectors. Later, both sets of feature vectors go through a Multi-Layer Perceptron (MLP) to match the feature sizes. Both the sets of transformed feature vectors are fed into a separate transformer encoder. Average pooling is applied on both sets of vectors, generating two fixed-size vectors representing the trajectory and the language command respectively. To predict whether they match or not, the two vectors are concatenated and fed into another MLP which in the end produces a scalar. Intuitively, through minimizing the cross-entropy loss, the two transformer encoders are encouraged to learn how the frames and the language command align to each other. In all the training in our experiments, the parameters of the image encoder and the language encoder are kept frozen.

The output of the Ext-LEARN framework which is the prediction probabilities are utilized to design intermediate rewards. The reward function is modelled as a function of both the past states and actions.

IV. EXPERIMENTAL EVALUATION

Similar to [8], we conducted experiments on the Atari game Montezuma's Revenge. In this game, the agent needs to navigate around different rooms with several obstacles such as ladders, ropes, doors, enemy objects, etc. With many objects and interactions, this game serves as a good environment to train an RL agent. Results from [8] serve as the baseline to compare the performance of our model.

Our model is trained and tested with the dataset obtained from [8]. Level 1 of Montezuma's revenge consists of 24 rooms. 14 are used for training, and 10 for validation and testing. Although objects remain the same across training and test/validation datasets, each room consists of only a subset of the objects arranged in various layouts. The training dataset is created with 160,000 (frame, language) pairs from the training set, and a validation dataset with 40,000 pairs from the validation set. A set of 15 tasks which involve the agent going from a particular start state to a fixed goal state across multiple rooms are considered. The agent interacts with various objects from a fixed set of objects during this period and for each task, on reaching the goal state, the agent receives a +1 reward from the environment and zero otherwise.

*Experiments*

In this experiment, we compare our new framework with two previous frameworks:
1. Ext-only: No language-based reward is used. Only a standard MDP with the original environment reward is considered.
2. Ext+Lang: The environment reward for successful completion of the task is combined with the language-based reward provided at each step.
3. New approach: The language reward from the proposed alignment model is used along with the environment reward.

The following metric are used to evaluate the performance of the model:
1. AUC: From each training, the number of timesteps is plotted against the number of successful episodes. The area under the curve (AUC) denotes how quickly the agent learns, and is used as the metric to compare two policy training runs.
2. Final Policy: To compare the final learned policy with ExtOnly, Ext+Lang and the new approach, policy evaluation is performed at the end of 500,000 training steps. For each policy training run, we use the learned policy for an additional 10,000 timesteps without updating it, and record the number of successful episodes.

## V. RESULTS

From the 10 policy learning test sets, we performed 10 policy learning runs for each set and each description. Fig. 3 below shows that our new policy is slightly faster than the Ext-learn and much faster than the Ext algorithm. It means that combining frames and languages together with complete action sequences is useful. The graphs show that the average number of successful episodes is around 1500 for the Ext-Lang algorithm, and this value is around 1760 for our algorithm. Also, our framework can reach 1500 successful episodes at around 420,000 time stamp which shows that it is around 15% faster than the Ext-learn algorithm.

Analysis of every test set was also performed. The correlation rate (variance) of each dataset is compared with the results in [8]. Fig. 4 and Fig. 5 shows the performance of the ExtOnly, Ext-Lang and proposed approach for task 4 and task 6. The proposed approach has a slightly higher success rate than the other baseline algorithms. Our algorithm has a small variance for simple tasks like task 4 and a large variance for more complex tasks such as task 6. Thus, our algorithm learns much faster than Ext-Lang and ExtOnly with average number of successful episodes about 1750 after 500,000 timesteps, while Ext+Lang is only at 1500 at that timestamp, which amounts to a 15% speed-up. Alternately, after 500,000 timesteps, ExtOnly is at around 800 average successful episodes, thereby giving a 90% relative improvement.

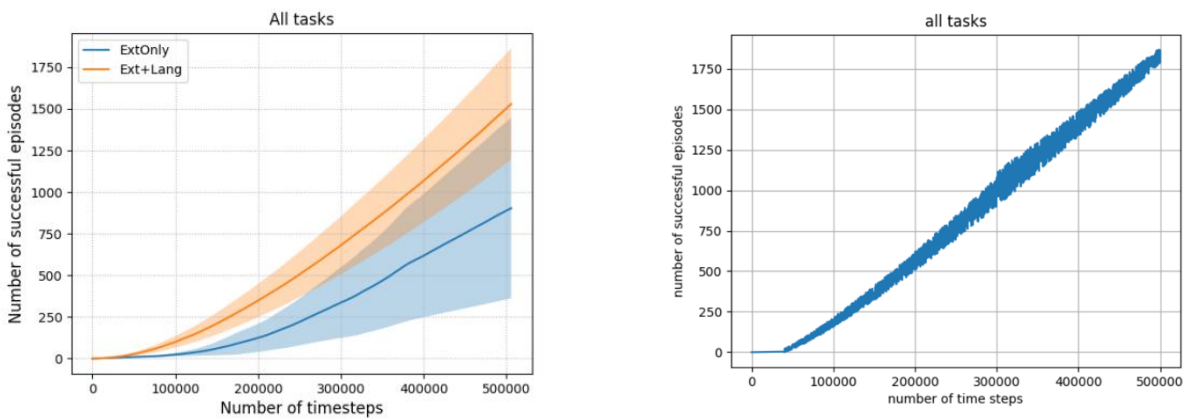

Fig. 3: Comparisons of the reward functions. (left) Solid lines denote the mean successful episodes averaged over all 15 tasks which reaches around 1500 for Ext-Lang and about 850 for ExtOnly. The shaded regions represent 95% confidence intervals. (right) The mean successful episodes averaged over all 15 tasks for the proposed model reaches 1760, outperforming ExtOnly and Ext-Lang.

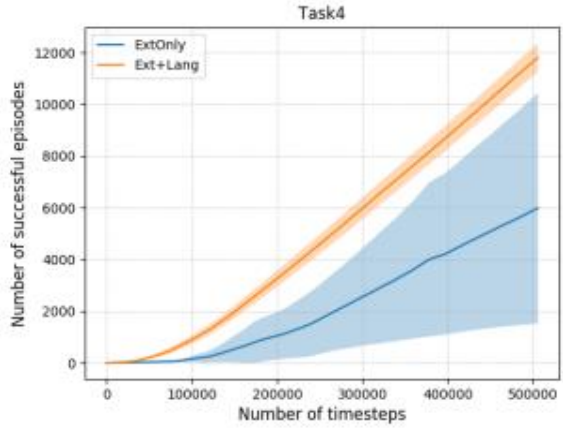 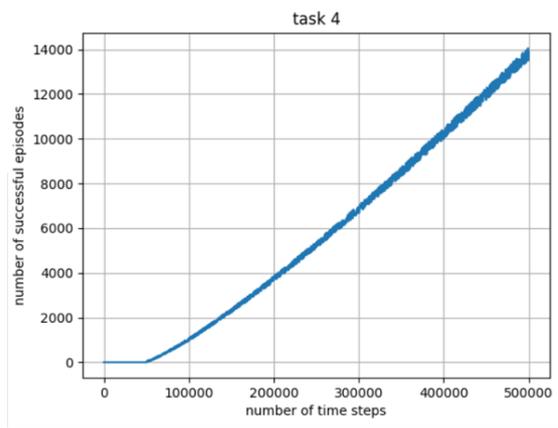

Fig. 4: Comparisons of the reward functions for task 4. (left) Solid lines denote the number of successful episodes for task 4 which reaches almost 12,00 for Ext-Lang and about 6000 for ExtOnly. The shaded regions represent 95% confidence intervals. (right) The number of successful episodes for task 4 for the proposed model reaches 14,000, outperforming ExtOnly and Ext-Lang.

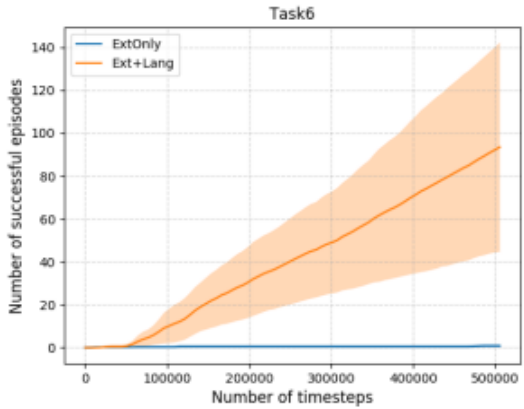 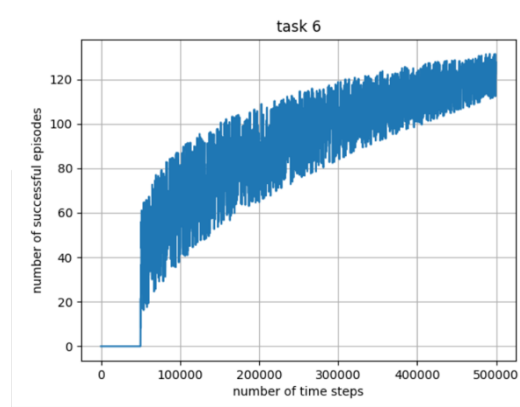

Fig. 5: Comparisons of the reward functions for a more complicated task 6. (left) Solid lines denote the number of successful episodes for task 6 which reaches almost 90 for Ext-Lang. The shaded region represents 95% confidence intervals. (right) The number of successful episodes for task 6 for the proposed model reaches 120, outperforming ExtOnly and Ext-Lang.

## VI. DISCUSSION

While utilizing natural language instructions for reward shaping has its benefits, there are some drawbacks to natural language-based rewards in RL. Firstly, grounding objects or actions within the environment using natural language proves to be challenging. For example, the instruction "Jump over the snake" needs grounding of object "snake" to the corresponding pixels in the image of the current state and "jump" must be mapped to the appropriate action in the action space. Additionally, natural language instructions could be ambiguous. "Jump over the snake" does not clearly specify the direction in which the agent should jump, leftward or rightward. Natural language also could be incomplete. The agent might have to move first in a certain direction before jumping over the snake. Synonyms of a word could be used to specify the same action ("move left" or "go left").

$MDP'$ considered includes only one-step language instruction $l$, while real-world scenarios usually involve multi-step instructions. A possible way to extend this work to handle multi-step instructions could be to determine if the language instruction is actually complete using another neural network-based or heuristic-based model. Then the prediction could be concatenated with the proposed model such that the model transitions to the new state only if the current instruction is completed.

The alignment model achieves an accuracy of 67% accuracy, whereas the LEARN model reached around 82%. This difference could be attributed to the separate supervised training process performed for grounding natural language commands in the environment in which the agent is deployed. Thus, the Ext-Learn model could be improved to provide better descriptions of the game trajectories which could lead to faster learning.

## VII. CONCLUSIONS

A natural language-based reward shaping model, which is an extension of the LEARN model, is proposed. Entire trajectories from the dataset are mapped to the corresponding language instructions representations obtained through BERT. This (trajectory, instruction) pair is fed into the Ext-LEARN model to predict if the instruction accurately describes the actions taken in the trajectory. This prediction probability is used to generate intermediate rewards in the language-aided RL model. The experiments performed show that using natural language-based rewards improves the policy learned and the time taken for the training process. Further, using a contextual word-embedding model such as BERT to generate vector representations of the natural language instructions and modelling the reward function based on complete action sequences improves the performance of the RL agent as compared to using GloVe embeddings with reward functions based only on past actions aggregated into action-frequency vectors.

APPENDIX

*A. Natural Language Instructions*

Table 1 provides some examples of natural language instructions collected from Amazon Mechanical Turk. These instructions do include variations in terms of vocabulary and length. There are spelling errors and semantically incomplete sentences. The model is still able to extract relevant information from such instructions.

Table 1: Natural Language Instructions corresponding the trajectories in the frames of the dataset.

| Sl. No. | Language Instructions |
|---|---|
| 1 | walk to the bars |
| 2 | climb up of the laddar and go the left |
| 3 | go to upstaird |
| 4 | RUN STRIGHT TOWRADS LADDER |
| 5 | two jumb while going left and take someone, then go left and right |

.